\documentclass{article}

\usepackage{arxiv}
\usepackage[utf8]{inputenc}
\usepackage[T1]{fontenc}
\usepackage{hyperref}
\usepackage{url}
\usepackage{booktabs}
\usepackage{amsfonts}
\usepackage{nicefrac}
\usepackage{microtype}
\usepackage{lipsum}
\usepackage{graphicx}
\usepackage{comment}

\graphicspath{{./images/}}

\title{A Survey On Enhancing Reinforcement Learning in Complex Environments: Insights from Human and LLM Feedback}

\author{
    Alireza Rashidi Laleh \\
    Cognitive Systems Lab. \\
    School of Electrical and Computer Engineering \\
    College of Engineering \\
    University of Tehran \\
    Tehran, Iran \\
    \texttt{alireza.rashidi16@ut.ac.ir} \\
    \And
    Majid Nili Ahmadabadi \\
    Cognitive Systems Lab. \\
    School of Electrical and Computer Engineering \\
    College of Engineering \\
    University of Tehran \\
    Tehran, Iran \\
    \texttt{mnili@ut.ac.ir} \\
}

\begin{document}

\maketitle

\begin{abstract}
Reinforcement learning (RL) is one of the active fields in machine learning, demonstrating remarkable potential in tackling real-world challenges. Despite its promising prospects, this methodology has encountered with issues and challenges, hindering it from achieving the best performance. In particular, these approaches lack decent performance when navigating environments and solving tasks with large observation space, often resulting in sample-inefficiency and prolonged learning times. This issue, commonly referred to as the curse of dimensionality, complicates decision-making for RL agents, necessitating a careful balance between attention and decision-making. RL agents, when augmented with human or large language models' (LLMs) feedback, may exhibit resilience and adaptability, leading to enhanced performance and accelerated learning. Such feedback, conveyed through various modalities or granularities including natural language, serves as a guide for RL agents, aiding them in discerning relevant environmental cues and optimizing decision-making processes. In this survey paper, we mainly focus on problems of two-folds: firstly, we focus on humans or an LLMs assistance, investigating the ways in which these entities may collaborate with the RL agent in order to foster optimal behavior and expedite learning; secondly, we delve into the research papers dedicated to addressing the intricacies of environments characterized by large observation space.
\end{abstract}


\section{Introduction} \label{sec:intro}
Reinforcement learning (RL) has emerged as a vital and dynamic area within machine learning, finding diverse applications in domains such as healthcare \cite{yu2021reinforcement} and natural language processing (NLP) \cite{uc2023survey}. In RL we have a sequential decision making problem in which, an agent is interacting with an environment-whether simulated or real-world-, collecting samples or experiences, and creating a behavior in order to achieve a presented task. Despite its many successes and advancements \cite{hambly2023recent}, RL still encounters fundamental challenges that hinder its widespread adoption and optimal performance \cite{casper2023open}.

Of these fundamental challenges we first have sample-inefficiency. The RL agents interact with the environment, perform exploration, and collect experiences in order to be able for informed decision-making, which leads to prolonged learning times for RL agents \cite{lin2020review}. The second challenge is the fact that the knowledge gained in the exploration phase by the RL agent may not necessarily be generalized to all or some unseen states and actions in the environment it is already interacting with \cite{packer2018assessing}. The third challenge deals with the fact that in complex environments with large observation space \cite{chiappa2024latent}, the RL agents need to perform a balance between decision-making and attention. In this type of environments, the RL agents are faced with the curse of dimensionality which is compounded in presence of the said challenges \cite{gosavi2009reinforcement}.

Among the proposed methodologies to address the challenges RL faces, there exist approaches that utilize external sources of information such as humans \cite{kaufmann2023survey} or large language models' (LLMs) \cite{pternea2024rl} feedback, alongside their own experiences and the knowledge gained from the environment \cite{bignold2023conceptual}. Humans or LLMs can provide feedback or presence for the RL agent in order for efficient and informed exploration and decision-making, expediting the RL agent's learning process and alleviating the challenges mentioned above. The feedback can take various forms including natural language, non-natural language, demonstrations, evaluative, and informative \cite{lin2020review, kaufmann2023survey}. Human or LLM provided natural language feedback is the more informative form of feedback, since, based on the level of their knowledge and awareness of the environment and task at hand, human or LLM can provide better feedback regardless of any hardships other forms experience \cite{jiang2019language}.

Overall, in this survey paper, our focus lies on two primary areas of literature. Firstly, in addition to the raw data received from the environment, we aim to supplement the RL agent with presence or feedback obtained from either human or an LLM. This feedback helps the RL agent to enhance performance, contributing valuable insights and guidance to aid its exploration and decision-making. These papers are available in Sections \ref{sec:humanf} and \ref{sec:llmf}, respectively. Secondly, we have addressed the curse of dimensionality in complex environments, where the observation space is large. In this challenge, the dilemma is the concurrency of attention and decision-making, since, it needs to mostly explore and experience many trials in order to learn and discern which parts of the environment are of importance to attend to for achieving optimal behavior and decision-making. These papers are available in Section \ref{sec:dmatt}.

The remainder of this paper is organized as follows: Section \ref{sec:back} provides fundamental concepts surrounding RL and LLMs. Sections \ref{sec:humanf} and \ref{sec:llmf} delve into RL with human feedback, and LLMs for RL, respectively. Section \ref{sec:dmatt} discusses the balance between decision-making and attention in environments with large observation spaces. Section \ref{sec:diss} underscores the importance of these topics. Finally, Section \ref{sec:conc} summarizes the findings and concludes the survey paper.

\section{RL with the Help of Human Feedback} \label{sec:humanf}
\begin{figure}[h]
  \centering
  \includegraphics[width=1\textwidth]{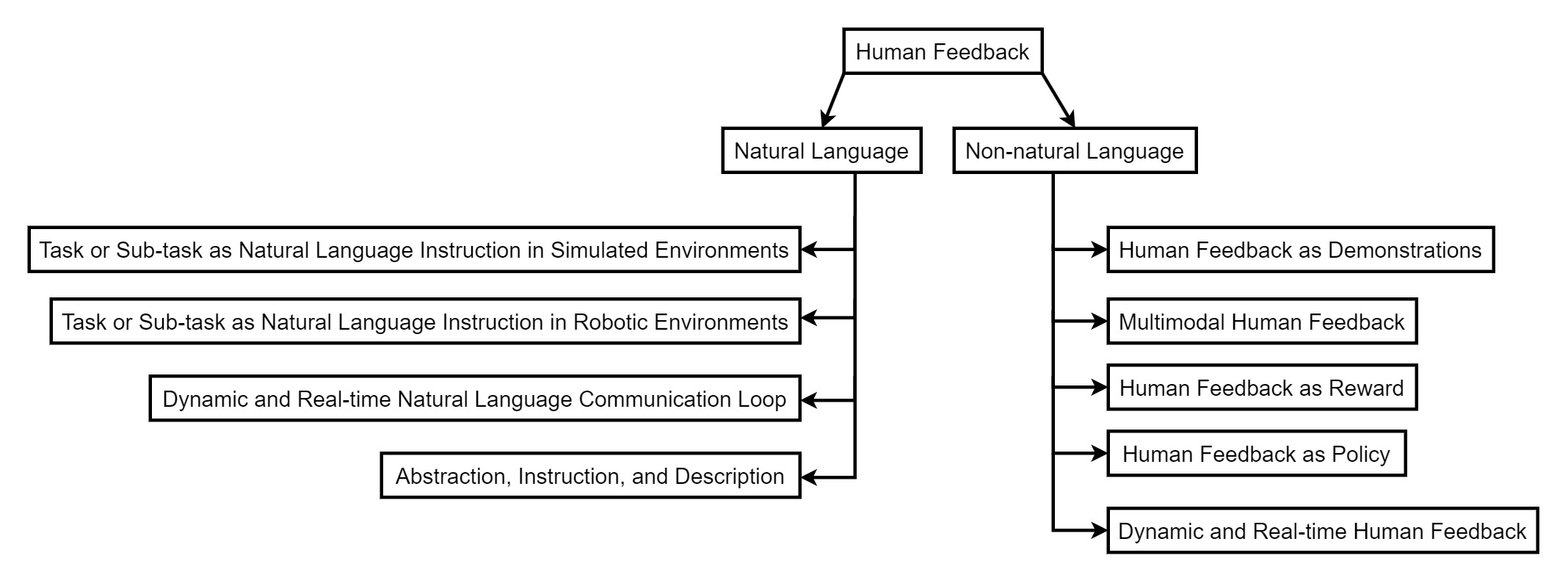}
  \caption{Hierarchy for Section \ref{sec:humanf}.}
  \label{fig:hierarchy-human}
\end{figure}

This section is mainly divided into two categories of research papers: first category consists of papers which utilize natural language as their feedback, and the second category consists of research papers which do not. Since, our main focus in this survey is on the first category, we first discuss these papers: sub-section \ref{sec:natlab}. Later in this section, we discuss papers which do not utilize such feedback/instruction: sub-section \ref{sec:nonnatlab}. Each category is also divided into further sub-sections, representing different clusters within each category. Figure \ref{fig:hierarchy-human} shows the taxonomy of the papers presented in this section.

\subsection{Natural Language Feedback/Instruction} \label{sec:natlab}
In this category, we discuss papers which utilize human natural language feedback. The feedback may be utilized in different forms or approaches by the RL agent. This category is divided into these sub-sections:

\begin{itemize}
    \item Section \ref{sec:natinssim}: papers which discuss natural language feedback/instruction in simulated RL environments.

    \item Section \ref{sec:natinsrob}: papers which are same as above item, but discuss in robotic and real-world environments, rather than simulated ones.

    \item Section \ref{sec:natloo}: papers which focus mainly on aspects in which, the human provides feedback/instruction in real-time and dynamically. This sub-section also discusses a line of communication between human and RL agent, in which the human provides corrective or evaluative feedback/instruction in natural language.

    \item Section \ref{sec:natabs}: papers which focus on different granularities of feedback, other than instruction itself, but are in natural language form. Meaning, the feedback may be a description or an abstraction, provided by human in natural language.
\end{itemize}

\subsubsection{Task or Sub-task as Natural Language Instruction in Simulated Environments} \label{sec:natinssim}
\begin{figure}[h]
  \centering
  \includegraphics[width=0.6\textwidth]{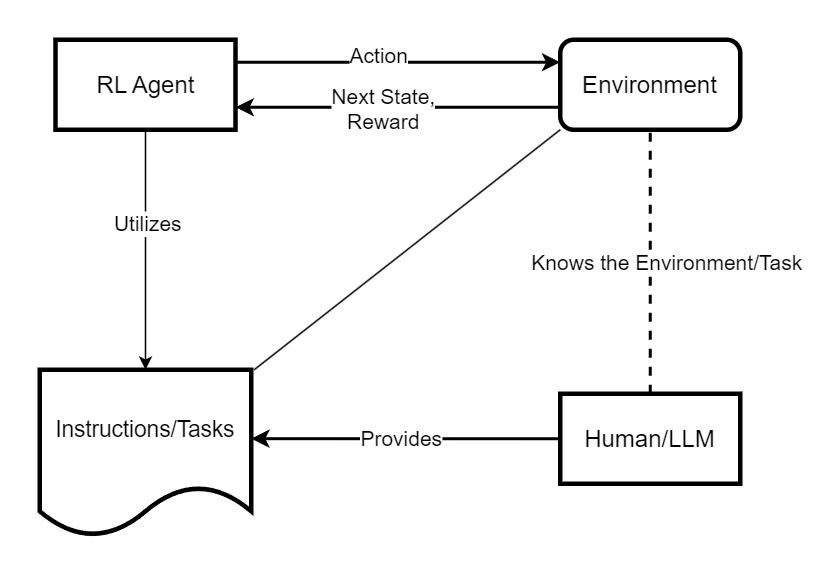}
  \caption{Abstract idea and architecture of papers in Sections \ref{sec:natinssim} and \ref{sec:natinsrob}.}
  \label{fig:fxy}
\end{figure}

This sub-section discusses papers which utilize natural language feedback/instruction as task or instruction to solve. The tasks or instructions have been provided by human before-hand and are presented for the RL agent to behave optimally. The human in this scenario is not involved dynamically or does not communicate in the learning process of RL agents. The abstract idea of these papers is shown in Figure \ref{fig:fxy}.

Kaplan et al. \cite{kaplan2017beating} introduce a deep RL approach in which, the RL agent is able to surpass previous benchmarks in Montezuma’s Revenge from the Atari Learning Environment \cite{bellemare2013arcade} with the help of natural language instructions. The RL agent receives both visual observations from the environment, and natural language instructions from humans. Combinations of these sources of information helps the RL agent to self-monitor its progress and grant rewards for completing instructions which represent sub-tasks alongside increasing game scores. 

Hill et al. \cite{hill2020human} utilize natural language instructions provided by humans to enhance and expedite the training of RL agents in environments. In their method, the RL agent utilizes pretrained language models such as BERT \cite{devlin2018bert}, to generate synthetic instructions, understand them, and behave accordingly. The authors have provided results showcasing the zero-shot transfer capabilities of the RL agents from synthetic instruction generated by BERT to human instructions in a 3D simulated environment suited for object detection and placement. 

Devo et al. \cite{devo2020deep} propose a method for visual navigation task in a 3D maze environment guided by natural language instructions. The RL agent receives natural language instructions and visual observations obtained by the environment for navigation. The RL agent is capable of interpreting the instructions to predict directions in order for improved navigation in the environment. 

Hermann et al. \cite{hermann2020learning} introduce a method which can perform navigation in real-world urban environments, called StreetNav, through images obtained from Google Street View. In their method, the RL agent is capable to interpret natural language driving instructions, which are provided by Google Maps. With the help of these driving instructions, and also visual information obtained from urban environments, the RL agent is able to reach any designated destination. 

Hu et al. \cite{hu2023language} introduce a framework for enhancing the performance of the RL agents in a multi-agent settings with the help of natural language instructions, and reaching equilibria during training faster. Humans express their desired policy through these natural language instructions. Based on these instructions and also observations from the environment, the proposed framework utilizes LLMs to generate a prior policy in order to help RL agents not diverge from human policies. 

Chen et al. \cite{chen2020ask} utilize human demonstrations annotated with human natural language instruction in a gridworld environment designed for crafting. They propose a two-layered hierarchical approach for multi-task and sparse reward setting. First layer is the low-level policy which is conditioned on the provided natural language instructions, and the second layer endeavors to interpret those instructions or generate new ones. With the help of provided dataset, the RL agent can perform imitation learning to achieve the presented tasks, or can learn to generate new instructions in novel tasks.

\subsubsection{Task or Sub-task as Natural Language Instruction in Robotic Environments} \label{sec:natinsrob}
This section is exactly same as \ref{sec:natinssim}, with one difference that, it discusses papers which are presented to tackle robotic environments and tasks. The abstract idea of these papers is shown in Figure \ref{fig:fxy}.

Bing et al. \cite{bing2023meta} utilize natural language instructions in robotic manipulation tasks to improve and expedite the training of deep RL approach. Their deep RL approach which utilizes meta-RL algorithm and language models, aims to reduce the time spent for trial-and-error learning during training of the RL agent, with the help of natural language instructions. The proposed approach is important since, it requires fewer trial-and-error steps for exploration and can understand tasks by interpreting natural language instructions and descriptions.

Stengel et al. \cite{stengel2022guiding} introduce a pretrained Transformer-based RL agent to utilize human natural language instructions in order to perform multi-step robotic manipulation tasks in a 3D environment. With the help of instructions and visual observation obtained by the environment, the RL agent is able to detect important locations for either grasping or placing of the objects. Since the location of objects is detected with the help of these inputs, the RL agent can perform tasks in fewer time-steps, thus, alleviating the problem of trial-and-error learning.

Lynch et al. \cite{lynch2023interactive} presents a framework to create a dynamic and real-time communication line between humans and robotic RL agents which can learn from natural language instructions. Their focus is mainly on the fact that the interaction between these two entities better to be real-time in order for the human to provide natural language instructions as tasks for the real-world robots. These robots can perform continuous control tasks such as robotic manipulations.

Sharma et al. \cite{sharma2022correcting} address the challenge of utilizing natural language corrections or instructions in planning objectives of the robot in order to enhance its performance and also clarify ambiguous human instructions. Their approach helps the RL agent to adjust and modify goals and its behavior dynamically, and also adapt to new goals based on the corrections human provides.

Shi et al. \cite{shi2024yell}, present a framework to enhance RL agent's performance in long-horizon robotic manipulation tasks with the help of real-time natural language human feedback. To achieve this, the authors propose a hierarchical policy framework that utilizes natural language instructions alongside low-level robotic control. At the high-level, the RL agent can generate language instructions autonomously, and human can provide corrective natural language feedback to further improve the performance. At the low-level control, the RL agent preforms action selection and behavior cloning based on interpreted natural language instructions.

\subsubsection{Dynamic and Real-time Natural Language Communication Loop} \label{sec:natloo}
\begin{figure}[h]
  \centering
  \includegraphics[width=0.6\textwidth]{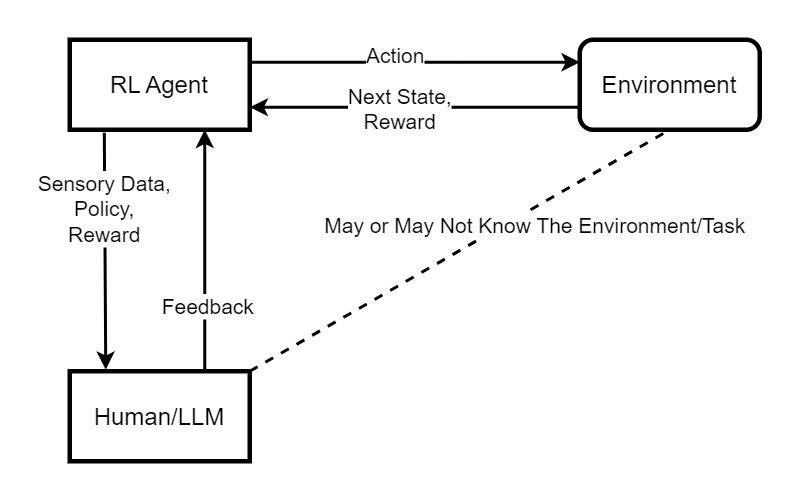}
  \caption{Abstract idea and architecture of papers in Section \ref{sec:natloo}.}
  \label{fig:sxy}
\end{figure}

This section discusses papers which utilize natural language feedback/instruction as task or instruction to solve. However, unlike Sections \ref{sec:natinssim} and \ref{sec:natinsrob}, the human in this scenario is involved dynamically in the learning process of the RL agent. Via a communication line, the human provides feedback for the RL agent at any time-step and the RL agent would behave accordingly. The abstract idea of these papers is shown in Figure \ref{fig:sxy}.

Surh et al. \cite{suhr2024continual} investigate the real-time natural language and binary feedback provided by the human for training of the RL agent. The authors emphasis on the fact that human can provide feedback at any time-step and these two entities collaborate together. Their appraoch consists of a contextual bandit setting, in which, the human provided feedback is converted into instant reward.

Wang et al. \cite{wang2022incorporating} incorporate natural language voice instructions in their approach to improve performance of the RL agent. Their setting is autonomous driving cars, in which the RL agent is controlling the car and the human provides voice feedback. In their approach, real-time speech-to-text methods and BERT-based classifier have been utilized to interpret human voice instructions and map them to proper and understandable actions for the RL agent. 

Van et al. \cite{van2022correct} address the challenge in which real-world RL agents might show incorrect behavior or any behavior that would result in failure. Based upon human natural language feedback, the RL agent can improve its action space, modify it, or remove some actions from it to prevent occurred failures. The authors have utilized non-expert humans to generate high-level and natural language feedback online and repair the RL agent's policy, rather than expert humans.

Tambwekar et al. \cite{tambwekar2023natural} introduce a framework in which humans can efficiently interpret or initialize RL agent's behavior or policy.  Inspired by decision tree algorithm, the authors in their paper endeavor to create an architecture in which the feedback is translated from natural language to a structured from such as lexical decision trees. Utilizing these, humans can efficiently utilize natural language descriptions to improve the RL agent's performance and policy. The RL agent utilizes such feedback and endeavors to reach the desired policy stated in natural language.

\subsubsection{Abstraction, Instruction, and Description} \label{sec:natabs}
\begin{figure}[h]
  \centering
  \includegraphics[width=0.6\textwidth]{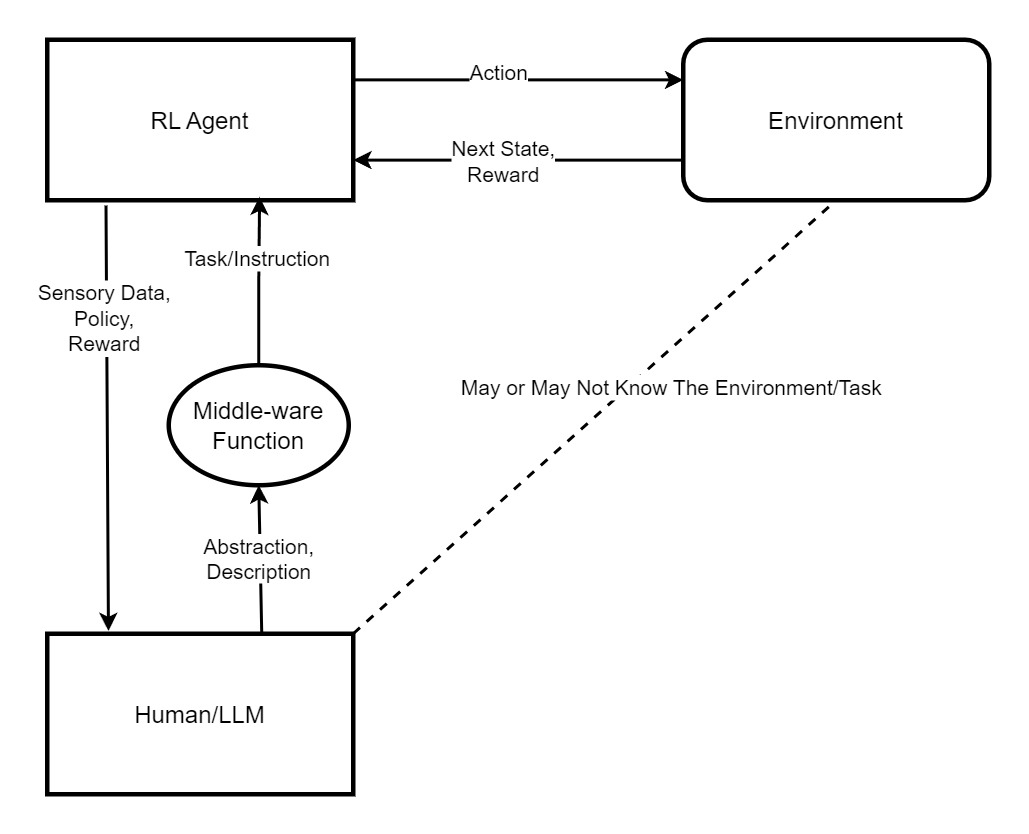}
  \caption{Abstract idea and architecture of papers in Section \ref{sec:natabs}.}
  \label{fig:kxy}
\end{figure}

This section discusses papers in which, the feedback can be of any type or granularity, such as a description, an instruction or an abstraction. Papers in this section mainly focus on the fact that the RL agent needs to be robust towards different types of natural language feedback, extracting necessary information useful for optimal behavior. The abstract idea of these papers is shown in Figure \ref{fig:kxy}.

Mirchandani et al. \cite{mirchandani2021ella} propose a framework for reward shaping which improves sample-efficiency of the RL agents in understanding and performing complex tasks defined in natural language. Their approach utilizes abstraction to guide exploration, believing that complex instructions can entail simpler ones. There are two parts in their approach: first, a classifier which classifies whether a low-level instruction has been done or not; second, a classifier which correlates the relevancy between low-level instructions with termination on high-level and complex tasks.

Sumers et al. \cite{sumers2022talk} propose an approach for RL agents to understand human preferences which have been presented in natural language. They utilize a contextual banding algorithm to distinguish between different granularities of human natural language feedback. First granularity is instructions and how they can provide information about policies. The second is descriptions and how they can provide information about reward functions. Integrating such mechanism in RL agents help them achieve autonomy in selection of granularity which can help them achieve better performance.

\subsection{Other Types of Feedback} \label{sec:nonnatlab}
In this category, we discuss papers which utilize human feedback which is not of natural language form. This category is divided into these sub-sections:

\begin{itemize}
    \item Section \ref{sec:nonnatloo}: papers which focus mainly on aspects in which, the human provides feedback in real-time and dynamically. This sub-section is similar to Section \ref{sec:natloo}, but with one difference and that is the feedback is not in natural language form.

    \item Section \ref{sec:nonnatmul}: papers which focus on feedback which is of different modalities; meaning, the RL agent needs to understand and utilize feedback from different modalities which are not necessarily of natural language form.

    \item Section \ref{sec:nonnatpol}: papers which discuss human feedback or involvement as policy. This means that the human suggests actions for the RL agent for optimal behavior.

    \item Section \ref{sec:nonnatrew}: papers which are same as above item, but discuss instead of utilizing the human feedback as a policy, we utilize human expertise as a reward function which provides rewards for any action chose by the RL agent.

    \item Section \ref{sec:nonnatdem}: papers which focus on human visual demonstrations as feedback. This type of papers is in line with inverse RL, however, with some minor differences.
\end{itemize}

\subsubsection{Dynamic and Real-time Human Feedback} \label{sec:nonnatloo}
This sub-section is similar to Section \ref{sec:natloo}. However, the humans do not provide natural language feedback and they utilize other forms such as evaluative feedback, preference-based, button clicking and etc. The abstract idea of these papers is similar to what is shown in Figure \ref{fig:sxy}.

Lou et al. \cite{luo2023human} address the challenge of sample inefficiency in RL by proposing a human-in-the-loop (HIL) framework. This framework incorporates human knowledge and expertise during the early stages of the RL agent's learning process to minimize the number of interactions with the environment. The RL agent requests human feedback only when it detects discrepancies in the Q-values computed by its policy.

Wu et al. \cite{wu2023toward} propose a framework to utilize human feedback and knowledge to enhance performance and training of the RL agent in autonomous driving setting. Humans can intervene or take control of the vehicle real-time during training and learning of the RL agent, leading to correcting failures of the RL agent. This leads to the RL agent learning optimal behavior facing situations in which it had failed before.

Moreira et al. \cite{moreira2020deep} explore the utilization of human interactive feedback in order to expedite the learning process of the deep RL agent. The authors investigate three methodologies in their paper: first one is autonomous RL agent itself without utilization of any feedback source, second one is the RL agent which utilizes feedback from other RL agent, and lastly the RL agent which utilizes human feedback in its architecture. The authors perform multiple simulated robotic experiments and showcase that utilizing feedback alongisde raw observations prove to be successful.

Li et al. \cite{li2023deploying} address the challenge of deploying RL agents which have been trained offline into real-world environments. The RL agent might have behaviors that are not fully optimal. Thus, the authors utilize human feedback in online settings to improve deployment process and further training of the RL agent. Human feedback is presented as finetuning process for the RL agent, and the RL agent endeavors to imrpove and modify its behaior in order to lessen the failures.

Guan et al. \cite{guan2020explanation} utilize human feedback in the form of saliency maps to provide binary evaluative feedback for improving the sample-inefficiency of the RL agent. By using these saliency maps alongside human binary feedback, humans can highlight regions in the observation space which are efficient to attend to for better decision-making. These regions can help the RL agent to expedite its learning process and lessen the trial-and-error process. 

Daniels et al. \cite{daniels2022expertise} address the challenge in which the humans' expertise in providing feedback for the RL agent are not completely correct or uniform, and there can exist noise or differences in some humans' feedback, which can lead to hindered performance of the RL agent. Humans' feedback are not usually uniform or similar regarding an RL agent, thus, this can lead to some complications in the process of helping the RL agent. The authors contribute to the fact that the RL agent needs to be able to select a specific human feedback based on their expertise, which can be the most helpful for it.

\subsubsection{Multimodal Human Feedback} \label{sec:nonnatmul}
In this sub-section, we discuss a paper which utilize different modalities for feedback such as, natural language, evaluative feedback, reward shaping and etc. These feedback modalities can be provided to the RL agent either simultaneously or sequentially. The RL agent needs to be compatible to any type of feedback and utilize it in its learning process. The abstract idea of these papers is shown in Figure \ref{fig:sxy}.

Trick et al. \cite{trick2022interactive} present an approach in which human feedback from multiple and different modalities are integrated into the RL agent. Their approach utilizes a Bayesian fusion method in order to combine the output distributions of base classifiers for each modality. This in turn highlights the confidence or uncertainty level of each modality and can help the RL agent for informed action-selection and also robustness towards the ambiguity of the feedback.

\subsubsection{Human Feedback as Policy} \label{sec:nonnatpol}
\begin{figure}[h]
  \centering
  \includegraphics[width=0.6\textwidth]{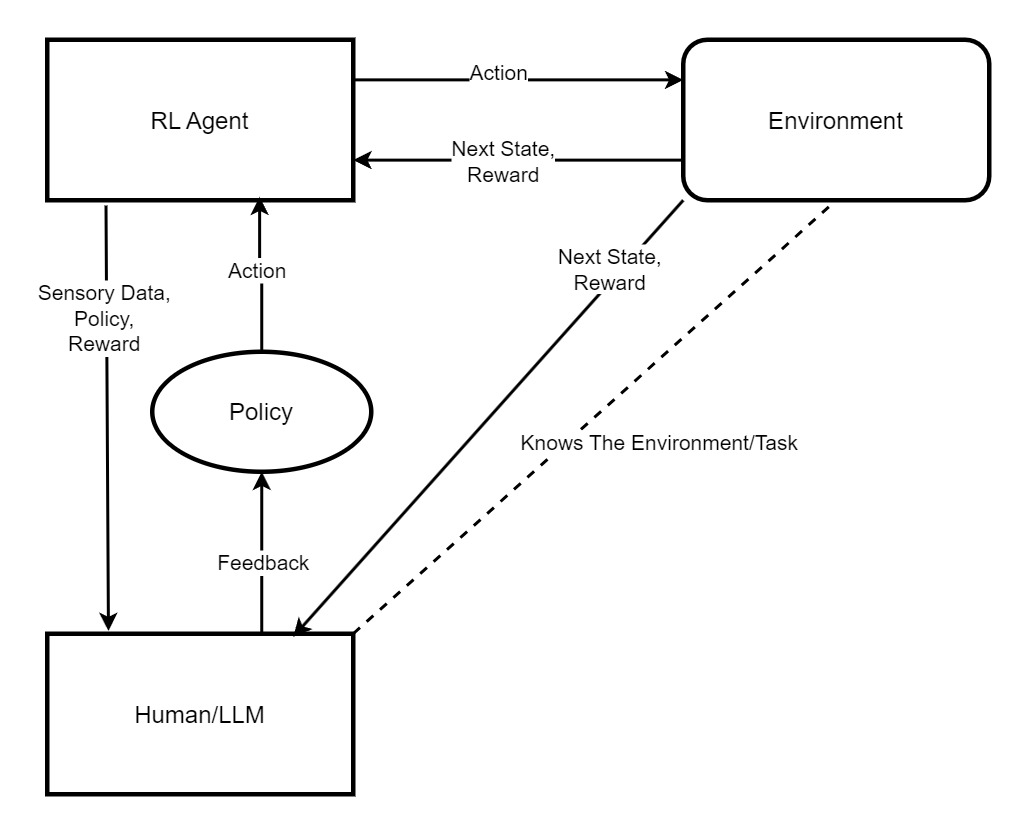}
  \caption{Abstract idea and architecture of papers in Sections \ref{sec:nonnatpol}.}
  \label{fig:zxy}
\end{figure}

In this sub-section, we discuss papers which utilize human selected or preferred actions as an independent policy. Humans either can suggest actions at any time-step, or they can intervene and modify action probabilities or values. The abstract idea of these papers is shown in Figure \ref{fig:zxy}.

Griffith et al. \cite{griffith2013policy} propose an approach to utilize human feedback as policy or action-selection alongside the RL agent's policy. In this approach the human provides feedback in form of a preferred action for the RL agent, which directly modifies the policy of the RL agent, instead of rewards or the RL agent's values. Due to its simplicity and adaptability, their approach can be utilized in real-world problems and scenarios.

Scherf et al. \cite{scherf2022learning} introduce an approach in which the RL agent is faced with human's inaccurate action feedback. In an interactive setting, the human can provide actions as feedback for the RL agent at any time-step, however, this feedback might prove to be incorrect. Thus, the RL agent needs evaluate this feedback from different aspects such as consistency, retrospective optimality, and behavioral cues. With the help of these aspects, the RL agent can build state-dependent trust in what the human has suggested or advised, and behave accordingly.

Frazier et al. \cite{frazier2019improving} investigate the efficacy of human selected actions as feedback in Minecraft game. The authors discuss the challenge of deep RL agents in such complex 3D environments and propose two interactive algorithms: Feedback Arbitration and Newtonian Action Advice. The authors discuss that these algorithms prove to be successful since, if the human provides feedback for the RL agent for multiple times, it would improve the RL agent's performance. Newtonian Action Advice can also prove to be useful whilst the RL agent is navigating the environment, wherein, the challenge of perceptual aliasing is important. Overall, the proposed algorithms are robust towards feedback with low levels of accuracy, since, the authors demonstrate that the high frequency of the provided feedback can make up for that.

\subsubsection{Human Feedback as Reward} \label{sec:nonnatrew}
\begin{figure}[h]
  \centering
  \includegraphics[width=0.6\textwidth]{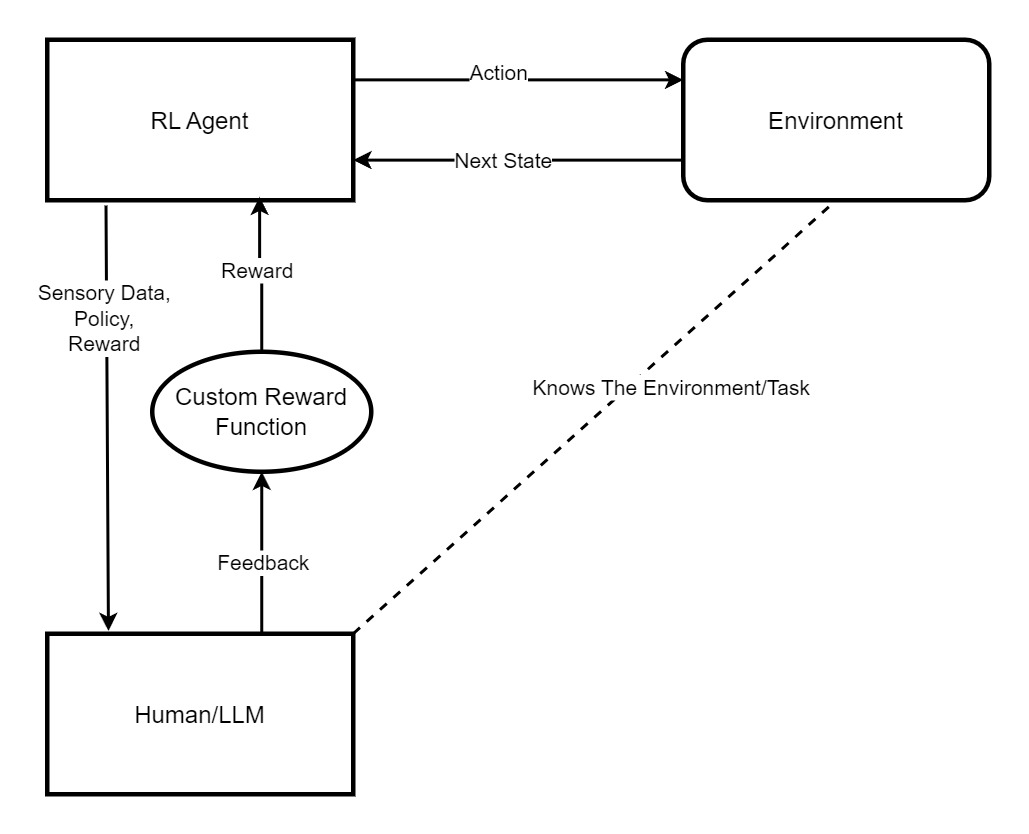}
  \caption{Abstract idea and architecture of papers in Section \ref{sec:natloo}.}
  \label{fig:txy}
\end{figure}

In this sub-section, we discuss papers which utilize human as a reward function. Humans may accept or reject selected behavior, or they can score chosen actions in certain time-steps. Thus, alongside environment reward, human feedback itself becomes a reward function. The abstract idea of these papers is shown in Figure \ref{fig:txy}.

Neider et al. \cite{neider2021advice} address the challenge of environments with sparse reward settings in RL. Their approach utilizes human feedback represented as deterministic finite automata (DFA). In their approach, a reward machine is iteratively learnt based on human feedback and also the observed rewards. Their approach guarantees optimal performance in long-horizon tasks, in turn reducing learning time.

Hejna et al. \cite{hejna2023few} present an approach in which human feedback is provided as reward function for the RL agent. The authors utilize human feedback or human preferences to pretrain reward models for wide range of tasks with different levels of diversity. This in turn causes the approach to be less reliable on the human presence in online and interactive setting, since the reward model is modelling human-presented preferences or rewards. Since, the RL agent is being trained on various tasks, their approach is inspired by meta-learning principles and multi-task learning settings.

Suay et al. \cite{suay2011effect} propose an approach in which the human can provide rewards as feedback for the RL agent real-time, in a real-world robotic setting. The provided human rewards are then integrated into the action-selection process of the RL agent. Since, the human is present in training process of the RL agent, it needs fewer trials and time-steps for exploration and finding optimal decision-making pattern, thus, proving this approach to be successful. 

\subsubsection{Human Feedback as Demonstrations} \label{sec:nonnatdem}
In this sub-section, we discuss papers which utilize human provided demonstrations as feedback, so that the RL agent can try to mimic the presented behavior. A term for these approaches is called "inverse RL". These papers focus on the fact that the RL agent would endeavor to learn a task by watching a human do it. The abstract idea of these papers is similar to what is shown in Figure \ref{fig:sxy}.

Li et al. \cite{li2018interactive} investigate utilizing human demonstrations and evaluative feedback to enhance the RL agents in real-world robotic settings. Their approach utilizes inverse RL principles in order for the RL agent to reach the optimal behavior and decision-making. The authors conclude that using their approach might not necessarily reach an optimal policy, however, it can produce decent and useful value functioon for the RL agent.

Huang et al. \cite{huang2024human} introduce an approach to deal with complex challenge of navigating mixed traffic environments for autonomous vehicles. They utilize human feedback which acts as a mentor to ensure safety and efficient traffic flow. Humans can intervene or guide the RL agents whenever needed for better and optimal behavior in uncertain or dangerous situations. Their approach does not require a custom reward fucntion and works with human demonstrations or interventions alone.

Lin et al. \cite{lin2023tag} present an approach in which human feedback is utilized for challenges of exploration and sample-inefficiency. They require the human to provide a few expert demonstrations for exploration process, then, they utilize a Gaussian process model to advise the RL agent actions with different levels of confidence. These demonstrations help the RL agent to learn how to perform efficient and more intentional exploration and decision-making process based on the human provided demonstrations, even in the parts of the environment it has not seen demonstration for.

\section{LLMs for RL} \label{sec:llmf}
\begin{figure}[h]
  \centering
  \includegraphics[width=1\textwidth]{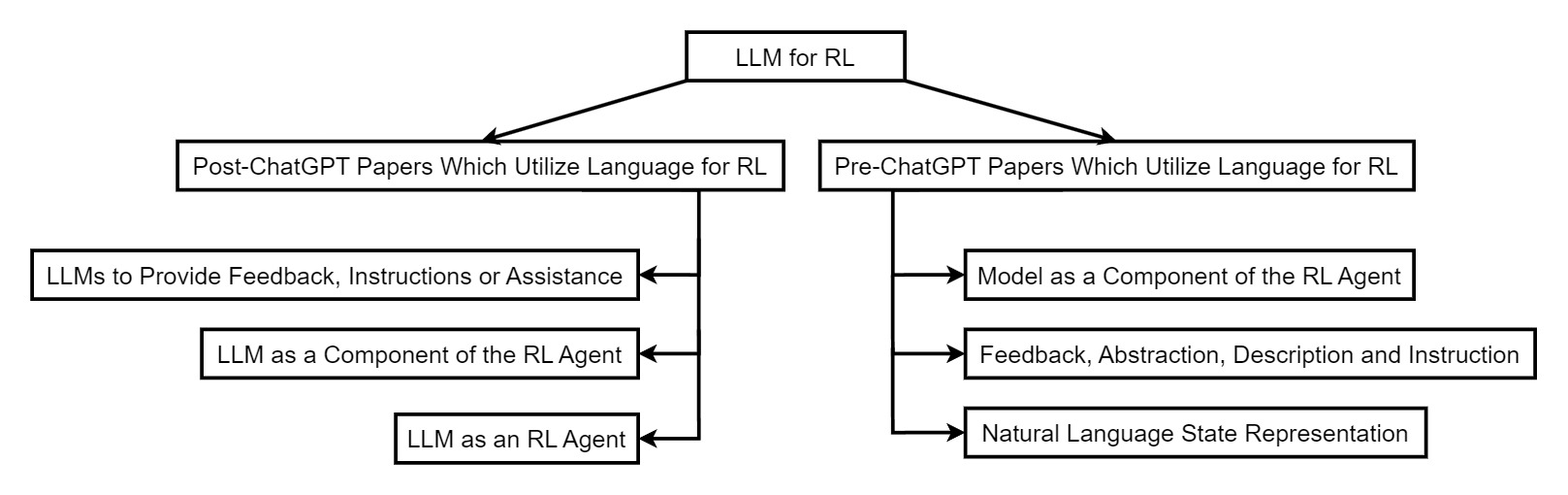}
  \caption{Hierarchy for Section \ref{sec:llmf}.}
  \label{fig:hierarchy-llm}
\end{figure}

This section is mainly divided into two categories of research papers: first category consists of papers from post-ChatGPT era which utilize LLM's assistance of any shape or form for the RL agent: Section \ref{sec:poscha}, and the second category consists of research papers which do the same, but are from pre-ChatGPT era and do not use any specific LLM in their approach: Section \ref{sec:precha}. Each category is divided into further sub-sections, representing different clusters within each category. Figure \ref{fig:hierarchy-llm} shows the taxonomy of the papers presented in this section.

\subsection{Post-ChatGPT Papers Which Utilize Language for RL} \label{sec:poscha}
In this category, we discuss papers which utilize LLM's natural language capabilities for RL agents. This category is divided into these sub-sections:

\begin{itemize}
    \item Section \ref{sec:llmfee}: papers which provide natural language feedback/instruction to the RL agent to achieve the task. These papers are similar to the papers from Sections \ref{sec:natabs} \ref{sec:natinsrob} and \ref{sec:natinsrob}.

    \item Section \ref{sec:llmcom}: papers which are same as above item, but instead of LLM being an extrinsic entity, it is a component of the RL agent. The LLM may also provide feedback in a real-time and dynamic setting unlike the above item.

    \item Section \ref{sec:llmage}: papers which focus mainly on scenarios in which, the LLMs can act as an RL agent. The LLM selects action based on a natural language high-level state representation created on top of the raw sensory data presented by the environment.
\end{itemize}

\subsubsection{LLMs to Provide Feedback, Instructions or Assistance} \label{sec:llmfee}
In this sub-section, we discuss papers primarily centered on LLMs providing natural language feedback/instructions to RL agents, often without direct human involvement. While some studies involve human prompters to elicit optimal feedback from LLMs, most focus solely on the LLM's role. The abstract idea of these papers is similar to what is shown in Figures \ref{fig:fxy}, \ref{fig:sxy} and \ref{fig:kxy}.

Du et al. \cite{du2023guiding} introduce a method in which pretrained LLMs provide instructions or goals for intrinsic motivation inside the RL agent for informed exploration in the environment. Upon completing each instruction by the RL agent, it receives a reward designed by the pretrained LLM, thus, lessening the effects of sparse reward setting. The LLM defines goals based on current state description the RL agent is in. LLM-generated goals are produced with the help of human prompter, and are also diverse in a sense that encourages the RL agent to perform meaningful exploration.

Barj et al. \cite{barj2024reinforcement} propose a method that utilizes LLMs' feedback to analyze and refine the RL agent's policy in situations where the RL agent has failed to generalize a specific goal to out-of-distribution environments. LLM's feedback acts as a reward model for the RL agent in order to identify potential failure scenarios.

Gu et al. \cite{gu2024mutual} investigate a cooperative teach-student learning framework in which the teache is the LLM, providing useful and abstract information for the RL agent. The RL agent acts as student to provide real-time information on its whereabouts to the LLM, and help the LLM in order to provide better feedback. This causes both entities to improve on themselves in order to accomplish mutual goals.

Karimpanal et al. \cite{karimpanal2023lagr} introduce a framework to utilize pretrained LLMs' natural language feedback to produce sets of decision-making behavior in order to expedite the learning of RL agent. Since, querying the LLM might be costy, the authors propose to utilize another RL agent to decide when to query the RL agent. Using the instructions and solutions produced through this sample-efficient querying system, the primary RL agent can perform optimal decision-making.

Kwon et al. \cite{kwon2023reward} explore the utilization of LLMs as reward functions in RL setting. In their approach the LLM is prompted with the examples of descriptions of observations and their corresponding optimal behavior and reward signals. Based on these examples in the prompts, the LLM is able to evaluate the behavior and decision-making of the RL agent towards any specific objective.

Hu et al. \cite{hu2023enabling} investigate the costly and time-consuming interactions between RL agents and LLMs in sequential decision-making tasks. The authors propose an approach in which the RL agent employs two policies, one of which is trained to decide when to query the LLM for natural language instructions to complete a given task. The pretrained LLM provides a planner role, in order for the RL agent to reach its goal. To accomplish this, one of the RL agent's policies is trained to realize when to query, and the other policy learns how to execute the instructions in the environment.

Liu et al. \cite{liu2022instruction} propose a transformer-based RL agent designed for natural language instruction-following tasks in vision-based environments. Their approach utilizes a pretrained multimodal transformer model capable of encoding and generating feature representations that integrate visual observations with natural language instructions.

Ma et al. \cite{ma2023eureka} introduce a method to utilize LLMs to autnomously design custom reward functions without any pre-defined prompting structure for any specific task. Their approach utilizes evolutionary optimization algorithms to utilize its capabilities in different settings such as zero-shot or few-shot learning. The authors demonstrate that the LLM in their approach is able to outperform human-designed reward functions.

\subsubsection{LLM as a Component of the RL Agent} \label{sec:llmcom}
\begin{figure}[h]
  \centering
  \includegraphics[width=0.6\textwidth]{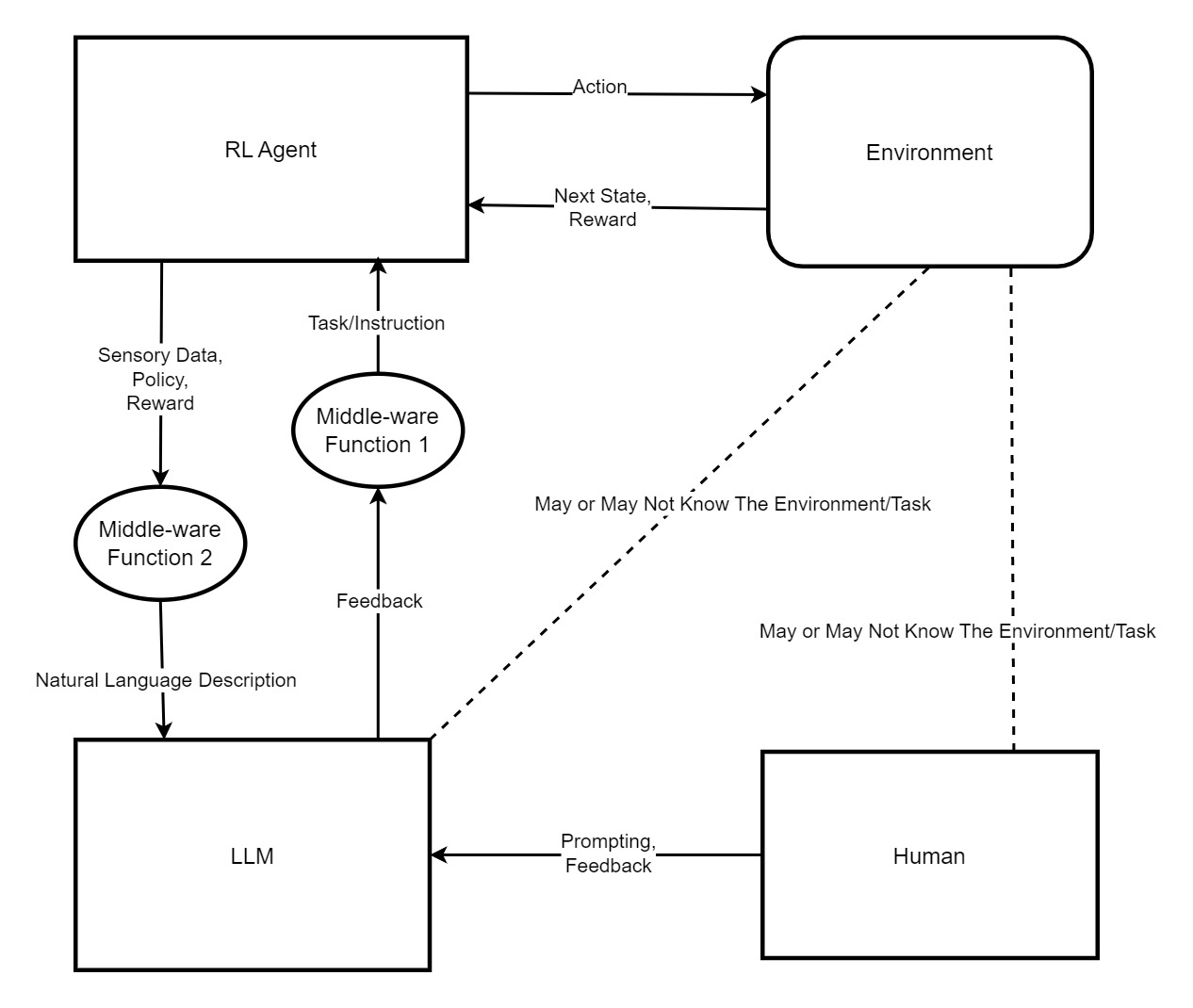}
  \caption{Abstract idea and architecture of papers in Sections \ref{sec:llmcom}.}
  \label{fig:pxy}
\end{figure}

In this sub-section, we discuss papers which investigate the integration of LLMs as intrinsic components of RL agents. Unlike the previous sub-section where LLMs provided feedback/instruction externally \ref{sec:llmfee}, here they are embedded within the RL agent framework, enabling real-time and dynamic interaction. This integration facilitates seamless communication between the RL agent and the embedded LLM, allowing both entities to access sensory data from the environment. The abstract idea of these papers is shown in Figure \ref{fig:pxy}.

Peng et al. \cite{peng2022inherently} introduce a framework in which the RL agent is able to explain its behavior or action-selection for performing tasks, whether online or post-hoc. Their approach is set for interactive text-based game environments. The authors utilize a symbolic knowledge graph state representation and a hierarchical graph attention mechanism for the RL agent's decision-making process. The provided are either immediate and local, or temporally extended post-hoc. The first type is generated by identifying key facts inside the knowledge graph whoch is influencing the decision-making process. The second type endeavors to include all the important segments of state space crucial in a trajectory for the decision-making.

Pan et al. \cite{pan2024hi} propose a hierarchical continual RL framework which utilizes LLMs for knowledge transfer. Their approach consists of two layers: a high-level policy formulation which is done by the LLM to generate a sequence of goals, and a low-level policy control to execute and train RL policies. These two layers provide feedback to each other and adjust themselves. For improvement of the performance of the RL agent, the high-level policy stores low-level policies in a skill library.

Klissarov et al. \cite{klissarov2023motif} introduce an approach to address the challenge of designing custom and intrinsic reward functions by utilizing the LLM's preferences over the RL agent's behavior. This intrinsic reward is combined with the reward received from the environment and utilized for the training of the RL agents. For each time-step, LLM sees a pair of performed action-selection or behavior by the RL agent and prefers one over the other.

Zhai et al. \cite{zhai2023building} integrate LLMs in RL agents to create embodied agents for human-AI interaction scenarios. The proposed framework of two stages. First, a pretained LLM is finetuned to map human natural language instructions into goals, and an RL policy endeavours to accomplish the stated goals. In the second stage, the authors train both the RL agent and the LLM together, so that these two entities are more aligned. In this optimization loop, the RL agent provides feedback for the LLM and causes the LLM to provide more informed feedback or set of goals for the RL agent. This leads to better understandment of the environment and also human provided instructions.

Prakash et al. \cite{prakash2023llm} present an approach in which the LLM is integrated into hierarchical RL agent in order to solve long-horizon tasks. The LLM guides the exploration of the RL agent in order to solve its sample-inefficiency. Selected actions or performed behavior by the RL agent is evaluated by the LLM based on the observed states and task description in order for biasing of the RL agent's decision-making process.

Quartey et al. \cite{quartey2023exploiting} propose utilizing LLMs in order to generate and learn auxiliary tasks alongisde the RL agent's main task in an object-centric environment. The authors aim to solve the sample-inefficiency of the RL agent. Their approach utilizes abstract temporal logic representationsof tasks and also object and context-aware embeddings obtained from the LLM to create the auxiliary tasks. These auxiliary tasks have the role of exploration for the RL agent so that it is able to accomplish the main task, therefore, enhancing the performance of the RL agent.

\subsubsection{LLM as an Agent} \label{sec:llmage}
\begin{figure}[h]
  \centering
  \includegraphics[width=0.6\textwidth]{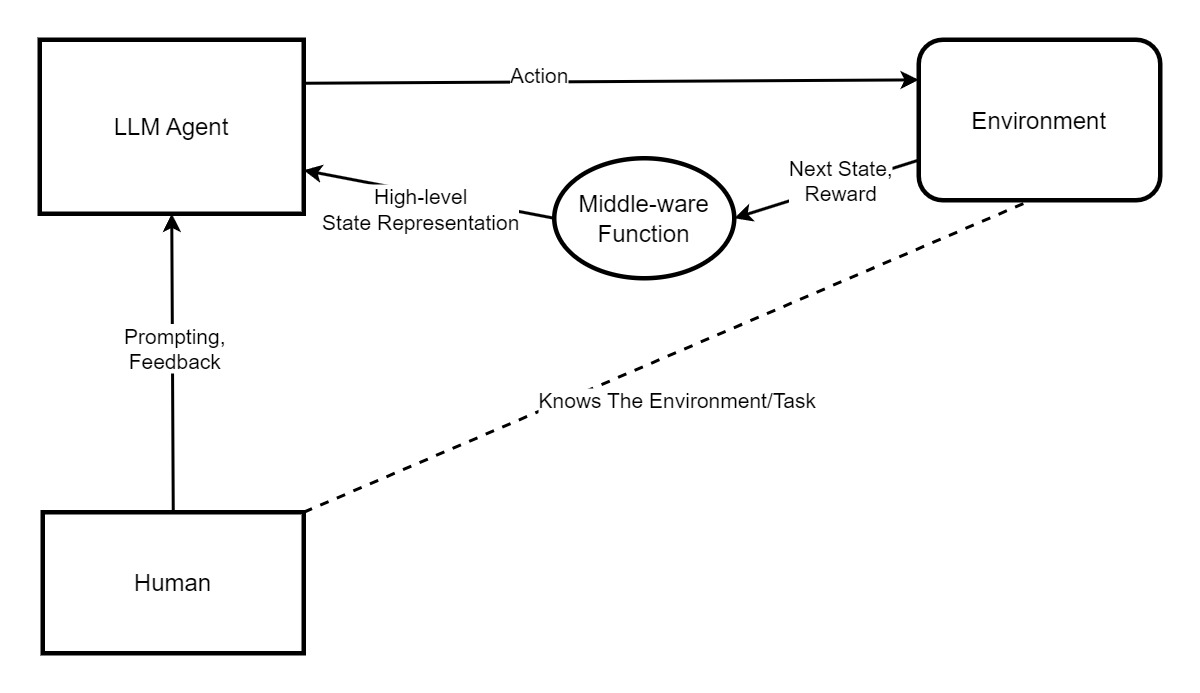}
  \caption{Abstract idea and architecture of papers in Section \ref{sec:llmage}.}
  \label{fig:rxy}
\end{figure}

In this sub-section, we discuss papers which explore the domain where LLMs serve as RL agents, leveraging their ability to interpret natural language high-level state representations derived from raw sensory data provided by the environment for decision-making. Notably, some LLMs possess multimodal capabilities, enabling them to process diverse data types, including visual observations. The abstract idea of these papers is shown in Figure \ref{fig:rxy}.

Tang et al. \cite{tang2024worldcoder} introduce a model-based agent in order to enable the LLM to program its knowledge gained from interactions with the environment in Python code. This agent learns the dynamics of the environment, reward and transition function, in order for improved planning and decision-making. This helps the agent to have an interpretable and sample-efficient learning process across gridworld environments.

Feng et al. \cite{feng2024natural} introduce an approach in which from the traditional RL and its key concepts related to it such as policies, value functions, task objectives, and etc. are translated to the natural language representation, and presented to the LLM. LLM endeavors to solve MDP and tabular tasks, which at the end leads to an interpretable behavior, easy-to-understand framework, efficient learning, diverse and capable reasoning without needing to human supervision. 

Yang et al. \cite{yang2023llm} introduce an approach in which LLMs act as an agent in 3D visual grounding robotic tasks. The utilized LLM does not require labeled training data in order to handle natural language queries for grounding objects. After understanding the provided query, the LLM engages with the provided grounding tool, and performs reasoning in order to recognize the spatial attributes of the objects for grounding and decision-making. LLMs can benefit from their commonsense knowledge in order for better understanding of the queries and spatial reasoning.

Sun et al. \cite{sun2023interactive} introduce an approach in which the LLM is utilized for interactive planning in partially observable MDP settings in robotics. The pretrained LLM agent faces uncertainty as it lacks complete observation or information at each time-step. This leads to the LLM endeavoring to father missing information from the environment in order to achieve understanding of its current state and then, informed decision-making. 

\subsection{Pre-ChatGPT Papers Which Utilize Language for RL} \label{sec:precha}
In this category, we discuss papers which utilize NLP-based and pre-ChatGPT models' natural language feedback/instruction. This category is divided into these sub-sections:

\begin{itemize}
    \item Section \ref{sec:prellmcom}: this sub-section is same as Section \ref{sec:llmcom}, with this difference that instead of utilizing LLMs, these papers utilize regular NLP-based models and architectures.

    \item Section \ref{sec:prellmfee}: this sub-section is same as Section \ref{sec:llmfee}, with this difference that instead of utilizing LLMs, these papers utilize regular NLP-based models and architectures.

    \item Section \ref{sec:prellmnat}: papers which focus on environments or tasks whose observation space is in natural language.
\end{itemize}

\subsubsection{Model as a Component of the RL Agent} \label{sec:prellmcom}
Papers in this sub-section are similar to the papers in Section \ref{sec:llmcom}. However, papers in this sub-section only utilize pre-ChatGPT models. The abstract idea of these papers is similar to what is shown in Figure \ref{fig:pxy}.

Hanjie et al. \cite{hanjie2021grounding} introduce a framework in which natural language descriptions are utilized within a multi-task environment for improved generalization capabilities. In their framework, the authors do not assume a prior mapping between textual descriptions and state observations. Instead, the RL agent needs to understand and recognize entities and informed behavior from the description based on reward signals. Their architecture utilizes attention to recognize and ground entities in a given textual description.

Wu et al. \cite{wu2017end} propose an approach in which the RL agent localizes objects in images based on natural language descriptions. The RL agent utilizes the spatial and temporal information obtained from the environment in order to modify and reshape bounding-boxes in order to localize the determined object. Their approach is robust towards scenarios in which there are many objects in the image, or objects having detailed attribution.

He et al. \cite{he2019read} propose an approach for grounding natural language descriptions in videos. The RL agent learns to efficiently recognize the accurate time-steps for beginning and ending of descriptions in video frames. The utilized policy is actor-critic, which iteratively observes the video frames and the description. Based on these observations, the RL agent learns to adjust beginning and ending points and doing so, it receives rewards which are generated from ground-trutth labels presented by humans for grounding.

\subsubsection{Feedback, Abstraction, Description and Instruction} \label{sec:prellmfee}
Papers in this sub-section are similar to the papers in Section \ref{sec:llmfee}. However, papers in this sub-section only utilize pre-ChatGPT models. The abstract idea of these papers is similar to what is shown in Figures \ref{fig:fxy}, \ref{fig:sxy} and \ref{fig:kxy}.

Shah et al. \cite{shah2018follownet} present an approach where the RL agent learns to navigate an environment using visual observations, depth data, and natural language instructions provided by a human. Based on the cross-attention mechanism between visual observations and textual instructions, the RL agent understands which parts of the environment are relevant for informed and better navigation and focus on them more. Since, the environment setting lacks dense rewards, the RL agent relies mostly on the cross-attention mechanism  and low-level control policies for navigation.

Eloff et al. \cite{eloff2021toward} explore the utilization of text-based natural language communication in collaborative multi-agent RL environments. This allows the RL agents to create collaboration between the RL agents without the need for predefined instructions. In their approach, one RL agent provide natural language instructions for others in order to navigate a maze-like environment. The sender RL agent encodes environment state and information and generates natural language instructions based on encoder-decoder architecture, and the receiver RL agent decodes those instructions into action sequences.

Jiang et al. \cite{jiang2019language} introduce a framework for hierarchical RL agents to utilize natural language as an abstraction. In their approach, two policies are employed: a high-level policy that generates natural language instructions and a low-level control policy that operates based on the high-level policy's guidance. The RL agent is able to reason in high-level space and consequently perform informed behavior. The authors also propose a method to relabel hindsight language instructions based on the alternative behavior the RL agent has performed for a specific instruction. This method helps the authors not to utilize reward shaping methods.

\subsubsection{Natural Language State Representation} \label{sec:prellmnat}
\begin{figure}[h]
  \centering
  \includegraphics[width=0.6\textwidth]{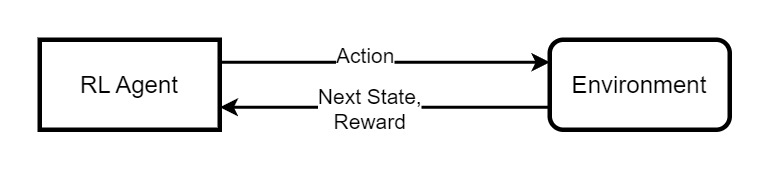}
  \caption{Abstract idea and architecture of papers in Sections \ref{sec:prellmnat}.}
  \label{fig:lxy}
\end{figure}

In this sub-section, we discuss papers which focus on environments or tasks driven by natural language, where both the state representation and available actions are expressed in language. These tasks often include text-based or narrative-driven games. The abstract idea of these papers is shown in Figure \ref{fig:lxy}.

Narasimhan et al. \cite{narasimhan2015language} propose an approach to learn control policies for text-based games, where state space, interactions, and action space are in natural language form. The presented game rewards are input to the RL agent as feedback for improved decision-making process. The authors utilize long short-term memory networks to encode textual representations into numerical vector representations which captures the game semantics.

Zhao et al. \cite{zhao2016towards} introduce a framework for task-oriented natural language dialog systems and also utilizing relational and structured databases. The RL agent has the language understanding capabilities to understand dialogues in order for informed decision-making and creation of strategic dialog policies. The authors also have utilized supervised learning techniques in their RL approach to expedite learning process.

\section{Trade-off between Decision-making and Attention} \label{sec:dmatt}
This section is mainly focused on research papers that either discuss complex environments which consist of large observation space, or have utilized some attention mechanism in their decision-making. This section is divided into these sub-sections:

In these papers, the authors have mostly utilized attention mechanisms in order to circumvent the challenges, and mainly focus on the trade-off of decision-making and attention.

\begin{itemize}
    \item Section \ref{sec:attnrl}: papers which discuss utilizing attention mechanism in their RL-based approach to solve challenges which consist of visual observations.

    \item Section \ref{sec:attcod}: papers which are almost same as above item, but discuss in environments whose observation space is large and suffer from curse of dimensionality.

    \item Section \ref{sec:attint}: papers which mainly focus on the interpretability and explainability of the produced attention masks between human subjects and RL agents.

    \item Section \ref{sec:attvln}: papers which specifically focus on the challenge of Visual-Language-Navigation.
\end{itemize}

\subsection{Attention-based RL for Visual Attention} \label{sec:attnrl}
In this sub-section, we explore how attention mechanisms are incorporated into RL frameworks to address challenges centered on visual observation space.

Borji et al. \cite{borji2010online} present an attention-based RL approach to learn object-based visual attention control in environments. Their model consists of three layers. Early visual processing layer performs bottom-up attention in which the RL agent detects objects based on highlighted locations in saliency maps. In higher visual processing, top-down attention is learnt via binary tree structure to identify the requested objects. Lastly, decision-making and learning delves into employing RL and provided information by previous layers to perform actions in attention-driven states.

Mayo et al. \cite{mayo2021visual} propose an attention-based visual navigation RL agent to combine semantic and spatial information for the goal object detection and finding. Based on the information the RL agent receives, the RL agent utilizes attention mechanism to make informed navigation behavior to recognize what objects exist in the environment and where they are located, in order to reach the desired goal object.

Barati et al. \cite{barati2019actor} introduce an RL approach for driving environments which consist of multiple views. Each view provides different sensory input about the car and also the environment, each have varying level of information and observability within them. By utilizing attention mechanism, the RL agent makes informed decisions regarding the importance of each view and how each can be beneficial. The RL approach utilizes actor-critic RL framework, and assigning a network to encode and create representations for each view. These representations are then integrated in a global neural network.

\subsection{Attention-based RL and Curse of Dimensionality} \label{sec:attcod}
In this sub-section, we explore papers tackling RL environments with large observation spaces, moving beyond visuals to include observations from other modalities. These papers tackle with the issue of curse of dimensionality, and mostly incorporate attention mechanisms into RL for efficient decision-making.

Pan et al. \cite{pan2023deep} introduce a deep RL approach which utilizes attention mechanism to solve the challenge of vehicle routing in real-world urban environments. Their approach is capable of performing learning and decision-making in partially-observable MDPs (POMDPs). Their approach also includes a graph-based POMDP model in order to observe and capture environment dynamics better. The deep RL's policy is adaptable to this information as well in order for enhanced learning performance.

Fei et al. \cite{fei2021optimizing} introduce a deep RL approach which endeavors to improve attention mechanism in NLP tasks. The authors have utilized three networks. The attention network which produces the attention weights for the environment states or in other words, the tokens. The other is about the policy and the process of decision-making, which endeavors to modify attention weights based on the environment states. Lastly, there is also a network to adjust the attention scores based on selected actions by the policy network, which receives rewards from the attention network. 

Reizinger et al. \cite{reizinger2020attention} explore environments which lack dense rewards. The proposed approach utilizes attention mechanism and curiosity-driven exploration for the RL agent, to enhance their learning performance. The attention mechanism emphasizes on parts of the relevant to both actor and critic networks of the RL agent. The attention mechanism emphasize the important parts and segments of the environment for the RL agent, and modifying its loss function in order to perform informed and improved curiosity-driven exploration for the RL agent.

Mott et al. \cite{mott2019towards} introduce an approach in which top-down attention mechanism is utilized for the RL agent to be able to focus on parts of the environment which are relevant and important for decision-making. The RL agent is actively interacting with the environment and by doing so, it is able to highlight most important parts. This enables the RL agent's decision-making process to be more interpretable and accessible. The RL agent is also able to adapt its architecture to novel states and planning future strategies.

\subsection{Interpretability of the Visual Attention Masks} \label{sec:attint}
In this sub-section, we review papers exploring the interpretability of attention mechanisms, connecting human understanding with RL agent decisions. They analyze attention masks to pinpoint key elements for decision-making and compare human and RL agent focus, revealing biases.

Leong et al. \cite{leong2017dynamic} investigate how RL agents with attention mechanism perform decision-making process in environments where the observations are of multi-dimension. They utilize human participants to perform experiments to determine which of these dimensions are relevant to the task and also the reward they receive. The authors utilize eye tracking tools and fMRI and let the human to engage in trial-and-error learning. The authors conclude their findnings by stating that the attention mechanism is important during learning, since, it biases value functions learnt and updated by the humans.

Zhang et al. \cite{zhang2020machine} propose to investigate how deep RL agents' visual attention mechanism performs in comparison with expert humans' attention while performing Atari games. The authors aim to examine the similarities between the visual representations generated by the two approaches and investigate how these representations impact the performance of RL agents. The authors investigate the saliency maps produced to gain insight into which parts of the environment has been of importance for humans and RL agents. Overall, the authors state that as time passes and RL agents gain experiences and knowledge, their attention model and produced visual representation becomes more similar to what the expert humans have produced.

\subsection{Vision-Language-Navigation} \label{sec:attvln}
In this sub-section, we discuss papers tackling the Visual-Language-Navigation (VLN) challenge, where RL agents must understand and follow navigation instructions given in natural language, adding complexity to decision-making. These papers aim to integrate language comprehension and spatial reasoning within RL frameworks, bridging the gap between textual instructions and visual perception.

Zitkovich et al. \cite{zitkovich2023rt} utilize a vision-language model which has been trained on large-scale internet data for robotic control tasks. Utilization of such data enhances generalization capabilities of the model and also helps it to acquire reasoning capabilities. The model is finetuned on robotic trajectory data, visual question answering, and also vision-language tasks so that the model is able to understand the given natural language task in order to accomplish the task. In their approach, the robotic actions which the model can take or perform are expressed as tokens, so that the mapping from instructions to a set of actions can be done easily.

He et al. \cite{he2023mlanet} introduce an approach for vision-language navigation task in which the RL agent performs navigation in a 3D environment based on visual observations and textual instructions. Their approach utilizes a multi-level architecture in which the raw textual instructions are segmented into sub-instructions in order for improved decision-making. Since, the complex raw textual instructions which may be long are not necessarily understandable for the RL agent, utilizing this method to segment instructions into sub-instructions can prove to be useful. This causes the RL agent to perform efficient decision-making process.

\section{Discussion} \label{sec:diss}
Thus far in this survey paper, we have discussed papers which focus on either the challenge of curse of dimensionality in RL, or assisting RL agents with the human or LLM feedback. Overall, there have been different methodologies explored and introduced. However, there still exists gaps and shortcomings in these areas. In this section we discuss few of such shortcomings that we think exist in literature and theorize possible solutions for them. These gaps and shortcomings are discussed in following sub-sections.

\subsection{Coexistence of Different Granularitites for Natural Language Feedback} \label{sec:coex}
Most papers which have utilized human or LLM natural language feedback, center around on providing only one granularity of feedback at the same time, and the provided feedback for RL agents is not multi-granularity. For example, the feedback has to be either an instruction or an abstraction, and these two types cannot be necessarily combined for further improvement of the performance. However, in some scenarios, the coexistence of different granularities of natural language feedback might prove to be successful. Overall, the RL agent must adeptly interpret and comprehend various forms of feedback and remain resilient, robust and adaptable to it, effectively integrating feedback into its processes. This becomes particularly critical in tasks with large observation spaces, necessitating numerous trial-and-error iterations to identify optimal strategies and behavior. 

\subsection{Autonomy of RL Agents}
Alongside using external source of information provided by either a human or an LLM, the RL agent needs to understand the task and be able to solve it itself. The RL agent is better not to be too dependent on provided feedback, rather, they have a choice to either accept, or reject it in incorporating them in their decision-making or exploration. This can go beyond simple accept/reject scenario, in which, the RL agent is able to extract information of any type of information: whether correct and helpful, or incorrect and unhelpful.

\subsection{Size of Descriptions and Instructions} 
Following the said shortcoming in Section \ref{sec:coex}, the provided natural language descriptions or instructions are usually not that very long, or in fewer steps at a time. This way, the reviewed papers are hindering the autonomy of RL agents, since, they are providing what exactly to attend to, or what exactly to do, in order for achieving the task goal. RL agents need to be robust towards any granularity and be able to extract useful information for decision-making and exploration.

\subsection{Human Awareness of the Tasks and Environments}
In most papers, the human necessarily knows the task, environment and goal of the scenario that the RL agent is facing with. The human may also know the optimal action and behavior in order to reach the goal. However, this may not be true for every challenge, task, environment or scenario, since, the human may not be familiar with either of these or the human may not know how to behave optimally. We can also mention the fact that the provided feedback needs to be correct, wherein some cases it might not. Since, humans may not always know the task, environment and they might know the optimal behavior and decision-making. 

\subsection{Dynamic Communication Line}
In most papers which utilize natural language feedback, there is no explicit dynamic line of communication for RL agents and humans/LLMs in which, both entities could communicate dynamically for better performance. In these papers, usually the natural language instructions or descriptions are given to the RL agent for improving the performance. Or that the humans/LLMs provide feedback only in certain time-steps. Lack of dynamism between these entities may prove not to be successful in scenarios where these two entities need to coexist for better decision-making of the RL agent.

\subsection{The Need for a Dataset}
In most papers which utilize natural language feedback and mostly are from pre-ChatGPT era, there is a need for a dataset to familiarize the RL agent to work with the task and the natural language feedback humans/LLMs provide. However, obtaining such dataset is cumbersome and needs domain or task expertise, and is not always available. This in turn may lead to less automation as human supervision and expertise is needed.

\section{Conclusion} \label{sec:conc}
In this survey paper we discuss papers which tackle the challenges reinforcement learning (RL) faces, such as sample-inefficiency, prolonged learning times, and curse of dimensionality or environments which have large observation space. The main methodology discussed in this survey paper is either using human/LLM's feedback and assistance, or learning and utilizing attention mechanism in their methodology. human or large language models (LLMs) feedback can be helpful for the RL agent for optimal behavior. Such feedback, conveyed through various modalities or granularities including natural language, serves as a guide for RL agents, aiding them in discerning relevant environmental cues and optimizing decision-making processes. On the other hand, by using attention mechanism the RL agent is able to attend to parts of the environment which is relevant for the optimal, informed, and faster decision-making process.

\section*{Acknowledgment}
The authors acknowledge the use of AI-based tools, including ChatGPT, for assisting in the preparation of this paper. These tools were utilized to create summaries of the cited works, as well as for text editing and enhancement, ensuring clarity and coherence throughout the manuscript.

\bibliographystyle{unsrt}  
\bibliography{references}

\begin{thebibliography}{10}

\bibitem{yu2021reinforcement}
Chao Yu, Jiming Liu, Shamim Nemati, and Guosheng Yin.
\newblock Reinforcement learning in healthcare: A survey.
\newblock {\em ACM Computing Surveys (CSUR)}, 55(1):1--36, 2021.

\bibitem{uc2023survey}
Victor Uc-Cetina, Nicolas Navarro-Guerrero, Anabel Martin-Gonzalez, Cornelius Weber, and Stefan Wermter.
\newblock Survey on reinforcement learning for language processing.
\newblock {\em Artificial Intelligence Review}, 56(2):1543--1575, 2023.

\bibitem{hambly2023recent}
Ben Hambly, Renyuan Xu, and Huining Yang.
\newblock Recent advances in reinforcement learning in finance.
\newblock {\em Mathematical Finance}, 33(3):437--503, 2023.

\bibitem{casper2023open}
Stephen Casper, Xander Davies, Claudia Shi, Thomas~Krendl Gilbert, J{\'e}r{\'e}my Scheurer, Javier Rando, Rachel Freedman, Tomasz Korbak, David Lindner, Pedro Freire, et~al.
\newblock Open problems and fundamental limitations of reinforcement learning from human feedback.
\newblock {\em arXiv preprint arXiv:2307.15217}, 2023.

\bibitem{lin2020review}
Jinying Lin, Zhen Ma, Randy Gomez, Keisuke Nakamura, Bo~He, and Guangliang Li.
\newblock A review on interactive reinforcement learning from human social feedback.
\newblock {\em IEEE Access}, 8:120757--120765, 2020.

\bibitem{packer2018assessing}
Charles Packer, Katelyn Gao, Jernej Kos, Philipp Kr{\"a}henb{\"u}hl, Vladlen Koltun, and Dawn Song.
\newblock Assessing generalization in deep reinforcement learning.
\newblock {\em arXiv preprint arXiv:1810.12282}, 2018.

\bibitem{chiappa2024latent}
Alberto~Silvio Chiappa, Alessandro Marin~Vargas, Ann Huang, and Alexander Mathis.
\newblock Latent exploration for reinforcement learning.
\newblock {\em Advances in Neural Information Processing Systems}, 36, 2024.

\bibitem{gosavi2009reinforcement}
Abhijit Gosavi.
\newblock Reinforcement learning: A tutorial survey and recent advances.
\newblock {\em INFORMS Journal on Computing}, 21(2):178--192, 2009.

\bibitem{kaufmann2023survey}
Timo Kaufmann, Paul Weng, Viktor Bengs, and Eyke H{\"u}llermeier.
\newblock A survey of reinforcement learning from human feedback.
\newblock {\em arXiv preprint arXiv:2312.14925}, 2023.

\bibitem{pternea2024rl}
Moschoula Pternea, Prerna Singh, Abir Chakraborty, Yagna Oruganti, Mirco Milletari, Sayli Bapat, and Kebei Jiang.
\newblock The rl/llm taxonomy tree: Reviewing synergies between reinforcement learning and large language models.
\newblock {\em arXiv preprint arXiv:2402.01874}, 2024.

\bibitem{bignold2023conceptual}
Adam Bignold, Francisco Cruz, Matthew~E Taylor, Tim Brys, Richard Dazeley, Peter Vamplew, and Cameron Foale.
\newblock A conceptual framework for externally-influenced agents: An assisted reinforcement learning review.
\newblock {\em Journal of Ambient Intelligence and Humanized Computing}, 14(4):3621--3644, 2023.

\bibitem{jiang2019language}
Yiding Jiang, Shixiang~Shane Gu, Kevin~P Murphy, and Chelsea Finn.
\newblock Language as an abstraction for hierarchical deep reinforcement learning.
\newblock {\em Advances in Neural Information Processing Systems}, 32, 2019.

\bibitem{kaplan2017beating}
Russell Kaplan, Christopher Sauer, and Alexander Sosa.
\newblock Beating atari with natural language guided reinforcement learning.
\newblock {\em arXiv preprint arXiv:1704.05539}, 2017.

\bibitem{bellemare2013arcade}
Marc~G Bellemare, Yavar Naddaf, Joel Veness, and Michael Bowling.
\newblock The arcade learning environment: An evaluation platform for general agents.
\newblock {\em Journal of Artificial Intelligence Research}, 47:253--279, 2013.

\bibitem{hill2020human}
Felix Hill, Sona Mokra, Nathaniel Wong, and Tim Harley.
\newblock Human instruction-following with deep reinforcement learning via transfer-learning from text.
\newblock {\em arXiv preprint arXiv:2005.09382}, 2020.

\bibitem{devlin2018bert}
Jacob Devlin, Ming-Wei Chang, Kenton Lee, and Kristina Toutanova.
\newblock Bert: Pre-training of deep bidirectional transformers for language understanding.
\newblock {\em arXiv preprint arXiv:1810.04805}, 2018.

\bibitem{devo2020deep}
Alessandro Devo, Gabriele Costante, and Paolo Valigi.
\newblock Deep reinforcement learning for instruction following visual navigation in 3d maze-like environments.
\newblock {\em IEEE Robotics and Automation Letters}, 5(2):1175--1182, 2020.

\bibitem{hermann2020learning}
Karl~Moritz Hermann, Mateusz Malinowski, Piotr Mirowski, Andras Banki-Horvath, Keith Anderson, and Raia Hadsell.
\newblock Learning to follow directions in street view.
\newblock In {\em Proceedings of the AAAI Conference on Artificial Intelligence}, volume~34, pages 11773--11781, 2020.

\bibitem{hu2023language}
Hengyuan Hu and Dorsa Sadigh.
\newblock Language instructed reinforcement learning for human-ai coordination.
\newblock {\em arXiv preprint arXiv:2304.07297}, 2023.

\bibitem{chen2020ask}
Valerie Chen, Abhinav Gupta, and Kenneth Marino.
\newblock Ask your humans: Using human instructions to improve generalization in reinforcement learning.
\newblock {\em arXiv preprint arXiv:2011.00517}, 2020.

\bibitem{bing2023meta}
Zhenshan Bing, Alexander Koch, Xiangtong Yao, Kai Huang, and Alois Knoll.
\newblock Meta-reinforcement learning via language instructions.
\newblock In {\em 2023 IEEE International Conference on Robotics and Automation (ICRA)}, pages 5985--5991. IEEE, 2023.

\bibitem{stengel2022guiding}
Elias Stengel-Eskin, Andrew Hundt, Zhuohong He, Aditya Murali, Nakul Gopalan, Matthew Gombolay, and Gregory Hager.
\newblock Guiding multi-step rearrangement tasks with natural language instructions.
\newblock In {\em Conference on Robot Learning}, pages 1486--1501. PMLR, 2022.

\bibitem{lynch2023interactive}
Corey Lynch, Ayzaan Wahid, Jonathan Tompson, Tianli Ding, James Betker, Robert Baruch, Travis Armstrong, and Pete Florence.
\newblock Interactive language: Talking to robots in real time.
\newblock {\em IEEE Robotics and Automation Letters}, 2023.

\bibitem{sharma2022correcting}
Pratyusha Sharma, Balakumar Sundaralingam, Valts Blukis, Chris Paxton, Tucker Hermans, Antonio Torralba, Jacob Andreas, and Dieter Fox.
\newblock Correcting robot plans with natural language feedback.
\newblock {\em arXiv preprint arXiv:2204.05186}, 2022.

\bibitem{shi2024yell}
Lucy~Xiaoyang Shi, Zheyuan Hu, Tony~Z Zhao, Archit Sharma, Karl Pertsch, Jianlan Luo, Sergey Levine, and Chelsea Finn.
\newblock Yell at your robot: Improving on-the-fly from language corrections.
\newblock {\em arXiv preprint arXiv:2403.12910}, 2024.

\bibitem{suhr2024continual}
Alane Suhr and Yoav Artzi.
\newblock Continual learning for instruction following from realtime feedback.
\newblock {\em Advances in Neural Information Processing Systems}, 36, 2024.

\bibitem{wang2022incorporating}
Mingze Wang, Ziyang Zhang, and Grace~Hui Yang.
\newblock Incorporating voice instructions in model-based reinforcement learning for self-driving cars.
\newblock {\em arXiv preprint arXiv:2206.10249}, 2022.

\bibitem{van2022correct}
Sanne Van~Waveren, Christian Pek, Jana Tumova, and Iolanda Leite.
\newblock Correct me if i'm wrong: Using non-experts to repair reinforcement learning policies.
\newblock In {\em 2022 17th ACM/IEEE International Conference on Human-Robot Interaction (HRI)}, pages 493--501. IEEE, 2022.

\bibitem{tambwekar2023natural}
Pradyumna Tambwekar, Andrew Silva, Nakul Gopalan, and Matthew Gombolay.
\newblock Natural language specification of reinforcement learning policies through differentiable decision trees.
\newblock {\em IEEE Robotics and Automation Letters}, 2023.

\bibitem{mirchandani2021ella}
Suvir Mirchandani, Siddharth Karamcheti, and Dorsa Sadigh.
\newblock Ella: Exploration through learned language abstraction.
\newblock {\em Advances in Neural Information Processing Systems}, 34:29529--29540, 2021.

\bibitem{sumers2022talk}
Theodore Sumers, Robert Hawkins, Mark~K Ho, Tom Griffiths, and Dylan Hadfield-Menell.
\newblock How to talk so ai will learn: Instructions, descriptions, and autonomy.
\newblock {\em Advances in Neural Information Processing Systems}, 35:34762--34775, 2022.

\bibitem{luo2023human}
Biao Luo, Zhengke Wu, Fei Zhou, and Bing-Chuan Wang.
\newblock Human-in-the-loop reinforcement learning in continuous-action space.
\newblock {\em IEEE Transactions on Neural Networks and Learning Systems}, 2023.

\bibitem{wu2023toward}
Jingda Wu, Zhiyu Huang, Zhongxu Hu, and Chen Lv.
\newblock Toward human-in-the-loop ai: Enhancing deep reinforcement learning via real-time human guidance for autonomous driving.
\newblock {\em Engineering}, 21:75--91, 2023.

\bibitem{moreira2020deep}
Ithan Moreira, Javier Rivas, Francisco Cruz, Richard Dazeley, Angel Ayala, and Bruno Fernandes.
\newblock Deep reinforcement learning with interactive feedback in a human--robot environment.
\newblock {\em Applied Sciences}, 10(16):5574, 2020.

\bibitem{li2023deploying}
Ziniu Li, Ke~Xu, Liu Liu, Lanqing Li, Deheng Ye, and Peilin Zhao.
\newblock Deploying offline reinforcement learning with human feedback.
\newblock {\em arXiv preprint arXiv:2303.07046}, 2023.

\bibitem{guan2020explanation}
Lin Guan, Mudit Verma, and Subbarao Kambhampati.
\newblock Explanation augmented feedback in human-in-the-loop reinforcement learning.
\newblock {\em arXiv preprint arXiv:2006.14804}, 2020.

\bibitem{daniels2022expertise}
Oliver Daniels-Koch and Rachel Freedman.
\newblock The expertise problem: Learning from specialized feedback.
\newblock {\em arXiv preprint arXiv:2211.06519}, 2022.

\bibitem{trick2022interactive}
Susanne Trick, Franziska Herbert, Constantin~A Rothkopf, and Dorothea Koert.
\newblock Interactive reinforcement learning with bayesian fusion of multimodal advice.
\newblock {\em IEEE Robotics and Automation Letters}, 7(3):7558--7565, 2022.

\bibitem{griffith2013policy}
Shane Griffith, Kaushik Subramanian, Jonathan Scholz, Charles~L Isbell, and Andrea~L Thomaz.
\newblock Policy shaping: Integrating human feedback with reinforcement learning.
\newblock {\em Advances in neural information processing systems}, 26, 2013.

\bibitem{scherf2022learning}
Lisa Scherf, Cigdem Turan, and Dorothea Koert.
\newblock Learning from unreliable human action advice in interactive reinforcement learning.
\newblock In {\em 2022 IEEE-RAS 21st International Conference on Humanoid Robots (Humanoids)}, pages 895--902. IEEE, 2022.

\bibitem{frazier2019improving}
Spencer Frazier and Mark Riedl.
\newblock Improving deep reinforcement learning in minecraft with action advice.
\newblock In {\em Proceedings of the AAAI conference on artificial intelligence and interactive digital entertainment}, volume~15, pages 146--152, 2019.

\bibitem{neider2021advice}
Daniel Neider, Jean-Raphael Gaglione, Ivan Gavran, Ufuk Topcu, Bo~Wu, and Zhe Xu.
\newblock Advice-guided reinforcement learning in a non-markovian environment.
\newblock In {\em Proceedings of the AAAI Conference on Artificial Intelligence}, volume~35, pages 9073--9080, 2021.

\bibitem{hejna2023few}
Donald~Joseph Hejna~III and Dorsa Sadigh.
\newblock Few-shot preference learning for human-in-the-loop rl.
\newblock In {\em Conference on Robot Learning}, pages 2014--2025. PMLR, 2023.

\bibitem{suay2011effect}
Halit~Bener Suay and Sonia Chernova.
\newblock Effect of human guidance and state space size on interactive reinforcement learning.
\newblock In {\em 2011 Ro-Man}, pages 1--6. IEEE, 2011.

\bibitem{li2018interactive}
Guangliang Li, Bo~He, Randy Gomez, and Keisuke Nakamura.
\newblock Interactive reinforcement learning from demonstration and human evaluative feedback.
\newblock In {\em 2018 27th IEEE International Symposium on Robot and Human Interactive Communication (RO-MAN)}, pages 1156--1162. IEEE, 2018.

\bibitem{huang2024human}
Zilin Huang, Zihao Sheng, Chengyuan Ma, and Sikai Chen.
\newblock Human as ai mentor: Enhanced human-in-the-loop reinforcement learning for safe and efficient autonomous driving.
\newblock {\em arXiv preprint arXiv:2401.03160}, 2024.

\bibitem{lin2023tag}
Ke~Lin, Duantengchuan Li, Yanjie Li, Shiyu Chen, Qi~Liu, Jianqi Gao, Yanrui Jin, and Liang Gong.
\newblock Tag: Teacher-advice mechanism with gaussian process for reinforcement learning.
\newblock {\em IEEE Transactions on Neural Networks and Learning Systems}, 2023.

\bibitem{du2023guiding}
Yuqing Du, Olivia Watkins, Zihan Wang, C{\'e}dric Colas, Trevor Darrell, Pieter Abbeel, Abhishek Gupta, and Jacob Andreas.
\newblock Guiding pretraining in reinforcement learning with large language models.
\newblock {\em arXiv preprint arXiv:2302.06692}, 2023.

\bibitem{barj2024reinforcement}
Houda Nait~El Barj and Th{\'e}ophile Sautory.
\newblock Reinforcement learning from llm feedback to counteract goal misgeneralization.
\newblock {\em arXiv preprint arXiv:2401.07181}, 2024.

\bibitem{gu2024mutual}
Shangding Gu.
\newblock Mutual enhancement of large language and reinforcement learning models through bi-directional feedback mechanisms: A case study.
\newblock {\em arXiv preprint arXiv:2401.06603}, 2024.

\bibitem{karimpanal2023lagr}
Thommen~George Karimpanal, Laknath~Buddhika Semage, Santu Rana, Hung Le, Truyen Tran, Sunil Gupta, and Svetha Venkatesh.
\newblock Lagr-seq: Language-guided reinforcement learning with sample-efficient querying.
\newblock {\em arXiv preprint arXiv:2308.13542}, 2023.

\bibitem{kwon2023reward}
Minae Kwon, Sang~Michael Xie, Kalesha Bullard, and Dorsa Sadigh.
\newblock Reward design with language models.
\newblock {\em arXiv preprint arXiv:2303.00001}, 2023.

\bibitem{hu2023enabling}
Bin Hu, Chenyang Zhao, Pu~Zhang, Zihao Zhou, Yuanhang Yang, Zenglin Xu, and Bin Liu.
\newblock Enabling intelligent interactions between an agent and an llm: A reinforcement learning approach.
\newblock {\em arXiv preprint arXiv:2306.03604}, 2023.

\bibitem{liu2022instruction}
Hao Liu, Lisa Lee, Kimin Lee, and Pieter Abbeel.
\newblock Instruction-following agents with multimodal transformer.
\newblock {\em arXiv preprint arXiv:2210.13431}, 2022.

\bibitem{ma2023eureka}
Yecheng~Jason Ma, William Liang, Guanzhi Wang, De-An Huang, Osbert Bastani, Dinesh Jayaraman, Yuke Zhu, Linxi Fan, and Anima Anandkumar.
\newblock Eureka: Human-level reward design via coding large language models.
\newblock {\em arXiv preprint arXiv:2310.12931}, 2023.

\bibitem{peng2022inherently}
Xiangyu Peng, Mark Riedl, and Prithviraj Ammanabrolu.
\newblock Inherently explainable reinforcement learning in natural language.
\newblock {\em Advances in Neural Information Processing Systems}, 35:16178--16190, 2022.

\bibitem{pan2024hi}
Chaofan Pan, Xin Yang, Hao Wang, Wei Wei, and Tianrui Li.
\newblock Hi-core: Hierarchical knowledge transfer for continual reinforcement learning.
\newblock {\em arXiv preprint arXiv:2401.15098}, 2024.

\bibitem{klissarov2023motif}
Martin Klissarov, Pierluca D'Oro, Shagun Sodhani, Roberta Raileanu, Pierre-Luc Bacon, Pascal Vincent, Amy Zhang, and Mikael Henaff.
\newblock Motif: Intrinsic motivation from artificial intelligence feedback.
\newblock {\em arXiv preprint arXiv:2310.00166}, 2023.

\bibitem{zhai2023building}
Shaopeng Zhai, Jie Wang, Tianyi Zhang, Fuxian Huang, Qi~Zhang, Ming Zhou, Jing Hou, and Yu~Liu.
\newblock Building open-ended embodied agent via language-policy bidirectional adaptation.
\newblock {\em arXiv preprint arXiv:2401.00006}, 2023.

\bibitem{prakash2023llm}
Bharat Prakash, Tim Oates, and Tinoosh Mohsenin.
\newblock Llm augmented hierarchical agents.
\newblock {\em arXiv preprint arXiv:2311.05596}, 2023.

\bibitem{quartey2023exploiting}
Benedict Quartey, Ankit Shah, and George Konidaris.
\newblock Exploiting contextual structure to generate useful auxiliary tasks.
\newblock {\em arXiv preprint arXiv:2303.05038}, 2023.

\bibitem{tang2024worldcoder}
Hao Tang, Darren Key, and Kevin Ellis.
\newblock Worldcoder, a model-based llm agent: Building world models by writing code and interacting with the environment.
\newblock {\em arXiv preprint arXiv:2402.12275}, 2024.

\bibitem{feng2024natural}
Xidong Feng, Ziyu Wan, Mengyue Yang, Ziyan Wang, Girish~A Koushiks, Yali Du, Ying Wen, and Jun Wang.
\newblock Natural language reinforcement learning.
\newblock {\em arXiv preprint arXiv:2402.07157}, 2024.

\bibitem{yang2023llm}
Jianing Yang, Xuweiyi Chen, Shengyi Qian, Nikhil Madaan, Madhavan Iyengar, David~F Fouhey, and Joyce Chai.
\newblock Llm-grounder: Open-vocabulary 3d visual grounding with large language model as an agent.
\newblock {\em arXiv preprint arXiv:2309.12311}, 2023.

\bibitem{sun2023interactive}
Lingfeng Sun, Devesh~K Jha, Chiori Hori, Siddarth Jain, Radu Corcodel, Xinghao Zhu, Masayoshi Tomizuka, and Diego Romeres.
\newblock Interactive planning using large language models for partially observable robotics tasks.
\newblock In {\em NeurIPS 2023 Workshop on Instruction Tuning and Instruction Following}, 2023.

\bibitem{hanjie2021grounding}
Austin~W Hanjie, Victor~Y Zhong, and Karthik Narasimhan.
\newblock Grounding language to entities and dynamics for generalization in reinforcement learning.
\newblock In {\em International Conference on Machine Learning}, pages 4051--4062. PMLR, 2021.

\bibitem{wu2017end}
Fan Wu, Zhongwen Xu, and Yi~Yang.
\newblock An end-to-end approach to natural language object retrieval via context-aware deep reinforcement learning.
\newblock {\em arXiv preprint arXiv:1703.07579}, 2017.

\bibitem{he2019read}
Dongliang He, Xiang Zhao, Jizhou Huang, Fu~Li, Xiao Liu, and Shilei Wen.
\newblock Read, watch, and move: Reinforcement learning for temporally grounding natural language descriptions in videos.
\newblock In {\em Proceedings of the AAAI Conference on Artificial Intelligence}, volume~33, pages 8393--8400, 2019.

\bibitem{shah2018follownet}
Pararth Shah, Marek Fiser, Aleksandra Faust, J~Chase Kew, and Dilek Hakkani-Tur.
\newblock Follownet: Robot navigation by following natural language directions with deep reinforcement learning.
\newblock {\em arXiv preprint arXiv:1805.06150}, 2018.

\bibitem{eloff2021toward}
Kevin~M Eloff and Herman~A Engelbrecht.
\newblock Toward collaborative reinforcement learning agents that communicate through text-based natural language.
\newblock In {\em 2021 Southern African Universities Power Engineering Conference/Robotics and Mechatronics/Pattern Recognition Association of South Africa (SAUPEC/RobMech/PRASA)}, pages 1--6. IEEE, 2021.

\bibitem{narasimhan2015language}
Karthik Narasimhan, Tejas Kulkarni, and Regina Barzilay.
\newblock Language understanding for text-based games using deep reinforcement learning.
\newblock {\em arXiv preprint arXiv:1506.08941}, 2015.

\bibitem{zhao2016towards}
Tiancheng Zhao and Maxine Eskenazi.
\newblock Towards end-to-end learning for dialog state tracking and management using deep reinforcement learning.
\newblock {\em arXiv preprint arXiv:1606.02560}, 2016.

\bibitem{borji2010online}
Ali Borji, Majid~Nili Ahmadabadi, Babak~Nadjar Araabi, and Mandana Hamidi.
\newblock Online learning of task-driven object-based visual attention control.
\newblock {\em Image and Vision Computing}, 28(7):1130--1145, 2010.

\bibitem{mayo2021visual}
Bar Mayo, Tamir Hazan, and Ayellet Tal.
\newblock Visual navigation with spatial attention.
\newblock In {\em Proceedings of the IEEE/CVF conference on computer vision and pattern recognition}, pages 16898--16907, 2021.

\bibitem{barati2019actor}
Elaheh Barati and Xuewen Chen.
\newblock An actor-critic-attention mechanism for deep reinforcement learning in multi-view environments.
\newblock {\em arXiv preprint arXiv:1907.09466}, 2019.

\bibitem{pan2023deep}
Weixu Pan and Shi~Qiang Liu.
\newblock Deep reinforcement learning for the dynamic and uncertain vehicle routing problem.
\newblock {\em Applied Intelligence}, 53(1):405--422, 2023.

\bibitem{fei2021optimizing}
Hao Fei, Yue Zhang, Yafeng Ren, and Donghong Ji.
\newblock Optimizing attention for sequence modeling via reinforcement learning.
\newblock {\em IEEE Transactions on Neural Networks and Learning Systems}, 33(8):3612--3621, 2021.

\bibitem{reizinger2020attention}
Patrik Reizinger and M{\'a}rton Szemenyei.
\newblock Attention-based curiosity-driven exploration in deep reinforcement learning.
\newblock In {\em ICASSP 2020-2020 IEEE International Conference on Acoustics, Speech and Signal Processing (ICASSP)}, pages 3542--3546. IEEE, 2020.

\bibitem{mott2019towards}
Alexander Mott, Daniel Zoran, Mike Chrzanowski, Daan Wierstra, and Danilo Jimenez~Rezende.
\newblock Towards interpretable reinforcement learning using attention augmented agents.
\newblock {\em Advances in neural information processing systems}, 32, 2019.

\bibitem{leong2017dynamic}
Yuan~Chang Leong, Angela Radulescu, Reka Daniel, Vivian DeWoskin, and Yael Niv.
\newblock Dynamic interaction between reinforcement learning and attention in multidimensional environments.
\newblock {\em Neuron}, 93(2):451--463, 2017.

\bibitem{zhang2020machine}
Ruohan Zhang, Sihang Guo, Bo~Liu, Yifeng Zhu, Mary Hayhoe, Dana Ballard, and Peter Stone.
\newblock Machine versus human attention in deep reinforcement learning tasks.
\newblock {\em arXiv preprint arXiv:2010.15942}, 2020.

\bibitem{zitkovich2023rt}
Brianna Zitkovich, Tianhe Yu, Sichun Xu, Peng Xu, Ted Xiao, Fei Xia, Jialin Wu, Paul Wohlhart, Stefan Welker, Ayzaan Wahid, et~al.
\newblock Rt-2: Vision-language-action models transfer web knowledge to robotic control.
\newblock In {\em Conference on Robot Learning}, pages 2165--2183. PMLR, 2023.

\bibitem{he2023mlanet}
Zongtao He, Liuyi Wang, Shu Li, Qingqing Yan, Chengju Liu, and Qijun Chen.
\newblock Mlanet: Multi-level attention network with sub-instruction for continuous vision-and-language navigation.
\newblock {\em arXiv preprint arXiv:2303.01396}, 2023.

\end{thebibliography}

\end{document}